\documentclass[letterpaper]{article}
\usepackage{natbib,alifeconf}
\usepackage{mdwlist,url}
\usepackage{mathptmx}
\usepackage{moresize}
\usepackage{stmaryrd}
\usepackage{amssymb}
\usepackage{graphicx}
\usepackage{verbatim}

\usepackage{accsupp} 

\newcommand*{\llbrace}{%
  \BeginAccSupp{method=hex,unicode,ActualText=2983}%
    \textnormal{\usefont{OMS}{lmr}{m}{n}\char102}%
    \mathchoice{\mkern-4.05mu}{\mkern-4.05mu}{\mkern-4.3mu}{\mkern-4.8mu}%
    \textnormal{\usefont{OMS}{lmr}{m}{n}\char106}%
  \EndAccSupp{}%
}
\newcommand*{\rrbrace}{%
  \BeginAccSupp{method=hex,unicode,ActualText=2984}%
    \textnormal{\usefont{OMS}{lmr}{m}{n}\char106}%
    \mathchoice{\mkern-4.05mu}{\mkern-4.05mu}{\mkern-4.3mu}{\mkern-4.8mu}%
    \textnormal{\usefont{OMS}{lmr}{m}{n}\char103}%
  \EndAccSupp{}%
}

\makeatletter
\newcommand{\shorteq}{%
  \settowidth{\@tempdima}{-}
  \resizebox{\@tempdima}{\height}{=}%
}

\usepackage{tikz}

\newcommand\mycirc[1]{%
  \tikz[baseline=(X.base)] 
    \node (X) [draw, shape=circle, inner sep=-1] {\strut {\small #1}};}
    
\usepackage[small,compact]{titlesec}

\makeatletter
\newcommand\npar{\@startsection{subsection}{2}{\z@}{-2\p@ \@plus -4\p@ \@minus -4\p@}{-0.5em \@plus -0.22em \@minus -0.1em}{\normalfontnormalsize\bfseries}}
\makeatother

\usepackage{calc} 

\newcommand*\phantomas[3][c]{%
\ifmmode 
\makebox[\widthof{$#2$}][#1]{$#3$}%
\else 
\makebox[\widthof{#2}][#1]{#3}%
\fi 
}

\newcommand{\kleisli}{ \:>\!=\!\!>}
\newcommand{\digest}{ \:> ! \!>\:}
\newcommand{\kleisliA} {\stackrel{A}{\kleisli}}

\newcommand{\join}{ -\!\!\!>\!\!\phantomas{@}{*}}
\newcommand{\eat}{ -\!\!\!>\!\!@}

\newcommand{\quit}{\phantomas{@}{*}-\!\!\!\!>}
\newcommand{\expel}{@-\!\!\!\!>}

\newcommand{\bind}{ \:>\!\!>\!=\:}
\newcommand{\bindA} {\stackrel{A}{\bind}}

\newcommand{\singleton}[1]{ \setl #1 \setr }
\newcommand{\nil} {\setl \setr}
\newcommand{\fail} {\supl \: \supr}

\newcommand{\range}{\;.\:.\;}

\newcommand{\funbox}[1]{\fbox{$ #1$}}

\newcommand{\cref}[1]{\overline{#1}}

\newcommand{\ceat}{{{\rm eat}}}
\newcommand{\cexpel}{{{\rm exit}}}
\newcommand{\cjoin}{{{\rm join}}}
\newcommand{\cquit}{{{\rm quit}}}
\newcommand{\csimilar}{{{\rm similar}}}
\newcommand{\cdissimilar}{{{\rm dissimilar}}}
\newcommand{\csame}{{{\rm same}}}
\newcommand{\cdifferent}{{{\rm different}}}
\newcommand{\cgrab}{{{\rm grab}}}
\newcommand{\cdrop}{{{\rm drop}}}
\newcommand{\cswap}{{{\rm swap}}}
\newcommand{\ccompose}{{{\rm compose}}}
\newcommand{\cdigest}{{{\rm digest}}}
\newcommand{\con}{{{\rm on}}}
\newcommand{\coff}{{{\rm off}}}
\newcommand{\cmakenext}{{{\rm bond}}}
\newcommand{\cmakeprev}{{{\rm bond}}\:'}
\newcommand{\ckillnext}{{{\rm unbond}}}
\newcommand{\ckillprev}{{{\rm unbond}}\:'}
\newcommand{\chands}{{{\rm hands}}}
\newcommand{\productrevs}{{{\rm prevs}}}
\newcommand{\cnexts}{{{\rm nexts}}}
\newcommand{\cbonds}{{{\rm bonds}}}
\newcommand{\cneighbors}{{{\rm neighbors}}}
\newcommand{\productarents}{{{\rm parents}}}
\newcommand{\ccontents}{{{\rm contents}}}
\newcommand{\cmembers}{{{\rm members}}}
\newcommand{\cothers}{{{\rm others}}}

\newcommand{\lreifiedobj} {\llbracket}
\newcommand{\rreifiedobj} {\rrbracket}
\newcommand{\lreifiedset} {\llbrace}
\newcommand{\rreifiedset} {\rrbrace}

\newcommand{\supl} {\langle}
\newcommand{\supr} {\rangle}
\newcommand{\setl} {\{}
\newcommand{\setr} {\}}
\newcommand{\stackl} {\left[}
\newcommand{\stackr}{\right]}
\newcommand{\nextbond}{>}

\title{Programs as Polypeptides}

\author{Lance R. Williams$^1$\\
\mbox{}\\
$^1$Department of Computer Science, University of New Mexico, Albuquerque, NM 87131\\ williams@cs.unm.edu}

\begin{document}
\maketitle
\begin{abstract}
We describe a visual programming language for defining behaviors manifested by reified actors in a
2D virtual world that can be compiled into programs comprised of sequences of combinators that
are themselves reified as actors.
This makes it possible to build programs that build programs from components of a few fixed types
delivered by diffusion using processes that resemble chemistry as much as computation.
\end{abstract}


\section{Introduction}

Self-replicating programs have been defined using computational models that vary in {\it expressiveness} and {\it verisimilitude}.
If we adopt the definition 
used in the field of programming languages \citep{felleisen},
then expressiveness varies along a spectrum that begins with {\it cellular automata} (CA) defined using lookup tables,
increases with {\it artificial chemistries} based on symbol rewrite rules,
and peaks in (more or less) conventional programming languages 
(which themselves vary along a spectrum that begins with machine language
and ends in high-level languages like Lisp).




By verisimilitude, we mean providing an interface with the affordances and limitations of a natural physics. 
Models with high verisimilitude define {\it virtual worlds}.
Because CAs are spatially embedded and governed by simple rules defined on local neighborhoods,
most would say that the verisimilitude of CAs is high.
However, since state is updated everywhere synchronously, and this (unlike a natural physics)
requires a global clock,
CAs are not {\it indefinitely scalable} \citep{bespoke}.
Because {\it asynchronous cellular automata} (ACA) do not suffer from this limitation
yet are just as powerful \citep{nakamura, berman, nehaniv},
ACAs are the gold standard in virtual worlds. 

Many artificial chemistries lack verisimilitude because the symbols that the rewrite rules transform are not embedded
in any physical space \citep{cham, paun, fontana}.
Others have far greater resemblance to real physical systems \citep{laing, turney, hutton}.
These assign symbols to positions in a virtual world, restrict interactions to local neighborhoods,
and rely on diffusion for data transport.

Programs written in conventional programming languages generally require a
{\it random access stored program} (RASP) computer to host them.\footnote{See \cite{williams14} for a notable exception.}
Because of {\it program-data equivalence}, RASPs permit relatively simple solutions to the self-replication problem based on {\it reflection.}
Yet self-replicating programs written in conventional programming languages are (in effect) stuck in boxes;
it makes no difference whether it is one big box \citep{ray} 
or many little boxes interacting in a virtual world \citep{adami1};
because they read, write, and reside in random access memories,
the programs themselves are fundamentally non-physical.


In the game of defining virtual worlds and creating self-replicating programs inside those worlds, there is a tradeoff
between the non-contingent complexity of {\it physical law} and the purely contingent complexity of the {\it initial conditions}
that define a program and its self-description.
We propose that the ratio of contingent and non-contingent complexity 
is positively correlated with the property that \cite{pattee} calls {\it semantic closure}.
Ideally, we would like to pursue an approach that combines the expressiveness of conventional programming languages
with the physical verisimilitude of ACAs while maximizing the ratio of contingent and non-contingent complexity.
To do this, we need to break programs out of their boxes;
we need reified programs that assemble copies of themselves from reified building blocks; 
we need to imagine {\it programs as polypeptides.}

Superficially, there is a similarity between the sequences of instructions that comprise
a machine language program and the
sequences of nucleotides and amino acids that comprise the
biologically important family of molecules known as {\it biopolymers}. 
It is tempting to view all of these sequences as `programs,'  broadly construed.
However, machine language 
programs and biopolymers differ in (at least)
one significant way, and that is the number of elementary building blocks
from which they are constructed.
The nucleotides that comprise DNA and RNA are only of four types;
the amino acids that comprise polypeptides are only of twenty;
and while bits might conceivably 
play the passive representational role of nucleotides,
they can not play the active functional role of amino acids;
this role can only be played by instructions.
While the instruction set of a simple RASP can be quite small, the number of distinct {\it operands}
that (in effect) modify the instructions is a function of the word size of the machine,
and is therefore (at a minimum) in the thousands.\footnote{Although they play many roles in machine language programs,
non-register operands are generally {\it addresses}.}
The implication for the study of self-replicating programs is profound: while biopolymers 
can be assembled by physical processes from building blocks
of a few fixed types, it is impossible to construct machine language programs for a RASP this way.

\begin{figure}[t]
\begin{center}
\includegraphics[scale = 0.32, trim = 10 0 0 0]{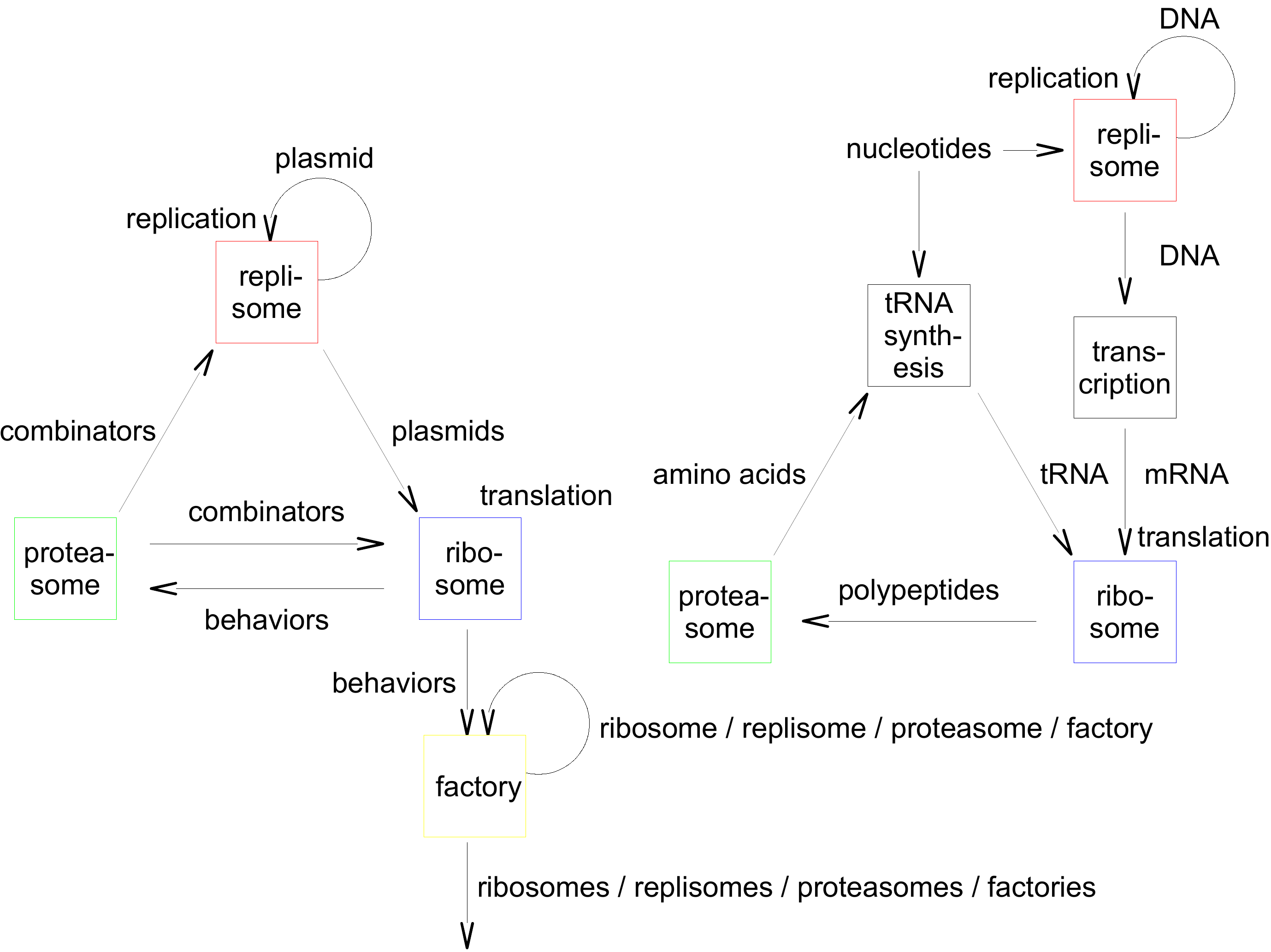}
\caption{Framework proposed in this paper (left).
Fundamental dogma of molecular biology (right).}
\label{dogma}
\end{center}
\end{figure}

DNA and RNA are copiable, transcribable and translatable descriptions of polypeptides.
DNA is (for the most part) chemically inert while polypeptides are chemically active.
Polypeptides can not serve as representations of themselves (or for that matter of anything at all) because their
enzymatic functions render this impossible. Information flows in one direction only.
\cite{watson} thought this idea so important that they called it ``the fundamental dogma of molecular biology.'' 
It is the antithesis of the program-data equivalence which makes reflection possible.
See Figure \ref{dogma}.

{\it Combinators} are functions with no free variables. 
In this paper we show how programs in a visual programming language just as expressive as machine language
can be compiled into sequences of combinators of only forty two types.
Where machine language programs would use iteration,
the programs that we compile into combinators employ {\it non-determinism}.
The paper culminates in the experimental demonstration of a {\it computational ribosome},
a `machine' in a 2D virtual world that assembles programs from combinators 
using inert descriptions of programs (also comprised of combinators)
as templates.

\section{Reified Actors}
Actors are created using three different constructors:
$[ \; ]^-$ creates {\it combinators},  $[ \; ]^+$ creates {\it behaviors}, and $[ \; ]_k$ creates  {\it objects.}
Like amino acids, which can be composed to form polypeptides, {\it primitive} combinators
can be composed to form {\it composite} combinators.
Behaviors are just combinators that have been repackaged with the $[ \; ]^+$ constructor.
Prior to repackaging, combinators do not manifest their function;
this might correspond (in our analogy) to the folding of a polypeptide chain into a {\it protein.}

Objects are containers that can contain other actors.
Each is one of four immutable types: $[ \; ]_0$, $[ \; ]_1$, $[ \; ]_2$ and $[ \; ]_3$.
For example, $[x,\: y,\: z\:]_2$ is an object of type two that contains three actors, $x$, $y$ and $z$.
Primitive combinators and empty objects have unit {\it mass.}
The mass of a composite combinator is the sum of the masses of the combinators of which it is composed.
The mass of an object is the sum of its own mass and the masses of the actors it contains.
Since actors can neither be created nor destroyed, mass is conserved.

Actors are reified by assigning them positions in a 2D virtual world.
Computations progress when actors interact with other actors in their
8-neighborhoods by means of the behaviors they manifest.
All actors are subject to {\it diffusion}.
An actor's diffusion constant decreases inversely with its mass.
This reflects the real cost of data transport in the (notional) ACA substrate.
Multiple actors can reside at a single site, but diffusion never moves
an actor to an adjacent occupied site if there is an adjacent empty site.

As with {\it membranes} in \cite{paun}, objects can be nested to any level of depth.
The object that contains an actor (with no intervening objects) is termed the actor's {\it parent.} 
An actor with no parent is a {\it root}.
Root actors (or actors with the same parent) can be associated with
one another by means of {\it groups} and {\it bonds.}
Association is useful because it allows working sets of actors to be constructed and
the elements of these working sets to be addressed in different ways.

The first way in which actors can associate is as members of a {\it group}.
All actors belong to exactly one group and this group can contain a single actor.
For this reason, groups define an {\it equivalence relation} on the set of actors.
A group of root actors is said to be {\it embedded}.
All of the actors in an embedded group diffuse as a unit and all behaviors manifested by actors
in an embedded group (or contained inside such an actor) share a finite time resource in a zero sum fashion.
Complex computations formulated in terms of large numbers of actors manifesting behaviors inside
a single object or group will therefore be correspondingly slow.
Furthermore, because of its large net mass, the object or group that contains them will also be
correspondingly immobile.

The second way in which actors can associate is by {\it bonding}.
Bonds are short relative addresses that are automatically updated as the actors they link undergo diffusion.
Because bonds are short ($L_1$ distance less than or equal to two),
they restrict the diffusion of the actors that possess them.
Undirected bonds are defined by the {\it hand} relation {\it H},
which is a {\it symmetric relation} on the set of actors, {\it i.e.,}
$H(x,y) = H(y,x)$.
Directed bonds are defined by the {\it previous} and {\it next} relations, {\it P} and {\it N},
which are {\it inverse relations} on the set of actors, 
{\it i.e.,} $P(x,y) = N(y,x)$.

If the types of combinators and behaviors were defined by the sequences of primitive combinators of which they are composed,
then determining type equivalence would be relatively expensive.
For this reason, we chose instead to define type using a simple recursive hash function
that assigns combinators with distinct multisets of components to distinct types:
the hash values of composite combinators are defined as the product of the hash values of their components;
primitive combinators have hash values equal to prime numbers.\footnote{We could instead use nested objects to label combinators so that they can be compared.
This would be like using codons constructed from nucleotides to label amino acids in transfer RNAs.}
Type equivalence for behaviors is defined in the same way, the types of combinators
and behaviors being distinct due to the use of different constructors.
Although this hash function is (clearly) not collision free, it is quite good
and it has an extremely useful property, namely, that composite combinators can be broken down (literally decomposed)
into their primitive components by prime factorization.\footnote{This is analogous to the function in the cell which is performed by the molecular assemblies
called proteasomes and in the organelles called lysosomes.}

Apart from composition, containment, group and bonds there is no other
mutable persistent state associated with actors.
In particular, there are no integer registers.
Primitive combinators exist for addressing individual actors or sets of actors using most of these relations.
These, and other primitive combinators for modifying actors' persistent states will be described later.

\section {Non-deterministic Comprehensions}
Sets can be converted into {\it superpositions} using the non-deterministic choice operator \citep{mccarthy}:
\begin{eqnarray*}
{\rm amb} \;  \nil & = & \fail\\
{\rm amb} \; \setl \: x, \: y \dots \setr & = & \supl \: x, \: y \dots \supr.
\end{eqnarray*}

\noindent When {\it amb} is applied to a non-empty set,  it causes
the branch of the non-deterministic computation that called {\it amb} to fork.
Conversely, empty sets cause the branch to fail.
When a branch fails, the deterministic implementation backtracks.

{\it Monads} are an 
abstract datatype that allows programmers to define rules for composing functions
that deviate from mathematically pure functions in prescribed ways.
Multivaluedness (represented by sets) and non-determinism (represented by superpositions) are just two examples.
The monad interface is defined by two operations called {\it unit} and {\it bind}.
{\it Unit} transforms ordinary values $a$ into {\it monadic values}, {\it e.g.,} 
${\rm unit}_A \; x = \left< x \right>$ where $A$ is the superposition monad.
Functions like {\it unit} that take ordinary values and return monadic values
are termed {\it monadic functions.}
{\it Bind} (the infix operator `$\bind$' in Haskell) allows monadic functions to be applied to monadic values.
This permits monadic functions to be chained;
the output of one provides the input to the next.

Monads are intimately related to set builder notation or {\it comprehensions}.
By way of illustration, consider the following non-deterministic comprehension that fails if $n$
is prime and returns a (non-specified) factor of $n$ if $n$ is composite:
\begin{eqnarray*}
\left< \; x \; | \; x \in \left<1 \range n-1\right>, \; y \in \left<1 \range x \right>, \; x \: y \; = \; n \; \right>.
\end{eqnarray*}
\cite{wadler92} showed that notation like the above is syntactic sugar for monadic expressions
and described a process for translating the former into the latter.
Comprehension {\it guards}, {\it e.g.,} $x \: y = n$, are translated using the function
\begin{eqnarray*}
{\rm guard}_M\;  \phantomas{\it False} {\it True}  & = & {\rm unit}_M \; \bot \\
{\rm guard}_M\;  {\it False} & = & {\rm zero}_M
\end{eqnarray*}
\noindent where $M$ is the monad and $\bot$ is {\it undefined.}
Because $zero_A$ is $\fail$, 
if ${\it guard}_A$ is applied to {\it False}, the branch of the computation 
that called ${\it guard}_A$ fails.
Conversely, if ${\it guard}_A$ is applied to {\it True}, the branch continues.
Using this device, the primality comprehension can be desugared as follows
\[
\lambda n \shortrightarrow ({\rm unit}_A \: n \bindA {\rm unit}_A \cdot (- \: 1) \bindA {\rm amb} \: \cdot \iota \bindA
\]
\[
\lambda x \shortrightarrow ({\rm unit}_A \: x \bindA {\rm amb} \: \cdot \: \iota \bindA {\rm unit}_A \cdot (\times \: x) \bindA
\]
\[
{\rm unit}_A \cdot (= \: n) \bindA {\rm guard}_A \bindA {\rm unit}_A \: x))
\]
\noindent where $(\iota \: x)$ equals $\setl 1 \! \range x \setr$.
\section {From Comprehensions to Dataflow Graphs}
Recall that our goal is to create programs comprised solely of combinators.
To maximize composability, these combinators should be of a single type,
yet the desugared comprehension above contains functions of many different types.
However, if sets are used to represent sets, singleton sets are used to represent scalars, 
and non-empty and empty sets are used to represent {\it True} and {\it False},
then the type signatures
\begin{eqnarray*}
\shortrightarrow \fbox{\phantomas{$g\:'$}{$f\:'$}} \shortrightarrow & :: & \setl a \setr \shortrightarrow \supl \setl a \setr \supr\\
_{\rightarrow}^{\rightarrow} \; \fbox{$g\:'$} \shortrightarrow & :: & \setl a \setr \shortrightarrow \setl a \setr \shortrightarrow \supl \setl a \setr \supr
\end{eqnarray*}
\noindent are general enough to represent the types of all functions in the desugared comprehension.
To prove this, we first show that {\it amb} can be lifted to the type, $\setl a \setr \shortrightarrow \supl \setl a \setr \supr$, as follows:
\begin{eqnarray*}
{\rm amb}\:' \;  \nil & = & \fail\\
{\rm amb}\:' \; \setl \: x, \: y \dots \setr & = & \supl\singleton{x}, \singleton{y} \dots \supr.
\end{eqnarray*}
\noindent We then devise a way to lift functions like $\iota$ with type, $a \shortrightarrow \setl a \setr$.
This is accomplished using the bind operator $(\bind_{\!\!\!\!S})$ for the set monad $S$.
The bind operator behaves like this
\[
\setl \: x,\: y \dots \: \setr \stackrel{S}{\bind} f  =  f \: x \: \cup f \: y \: \cup \dots
\]
\noindent and can be defined as follows
\[
(\stackrel{S}{\bind} f) = {\rm join}_S \cdot \; ({\rm map}_S \: f)
\]
\noindent where ${\it join}_S$ is right fold of $(\cup)$ and
\[
{\rm map}_S \; f \; \setl \: x,\: y \dots \: \setr =  \setl f \: x, \: f \: y \: \dots \setr.
\]
Bind can then be used with ${\it unit}_A$ to lift $\iota$ into a function
\[
\iota\:' = {\rm unit}_A \cdot (\stackrel{S}{\bind} \iota)
\]
\noindent with the type, $\setl a \setr \shortrightarrow \supl \setl a \setr \supr$,
as demonstrated below
\begin{eqnarray*}
\iota \:' \setl \: x,\: y \dots \: \setr  & = & \supl \: \iota \: x \: \cup \iota \: y \: \cup \dots \supr.
\end{eqnarray*}
\noindent Next we define two functions of type, $\setl a \setr \shortrightarrow \supl \setl a \setr \supr$, to replace {\it guard}.
The first causes a computation to fail when its argument is empty while the second does the opposite:
\begin{eqnarray*}
{\rm some}\:' \; \nil & = & \fail\\
{\rm some}\:' \; \setl \: x, \: y \dots \: \setr & = &  \supl \setl \: x, \: y \dots \: \setr \supr \\
{\rm none}\:' \;  \nil & = & \supl \setl \setr \supr\\
{\rm none}\:' \; \empty \setl \: x, \: y \dots \: \setr & = &  \fail.
\end{eqnarray*}

\begin{figure}[t]
\begin{center}
\includegraphics[scale = 0.3, trim = 0 300 0 0]{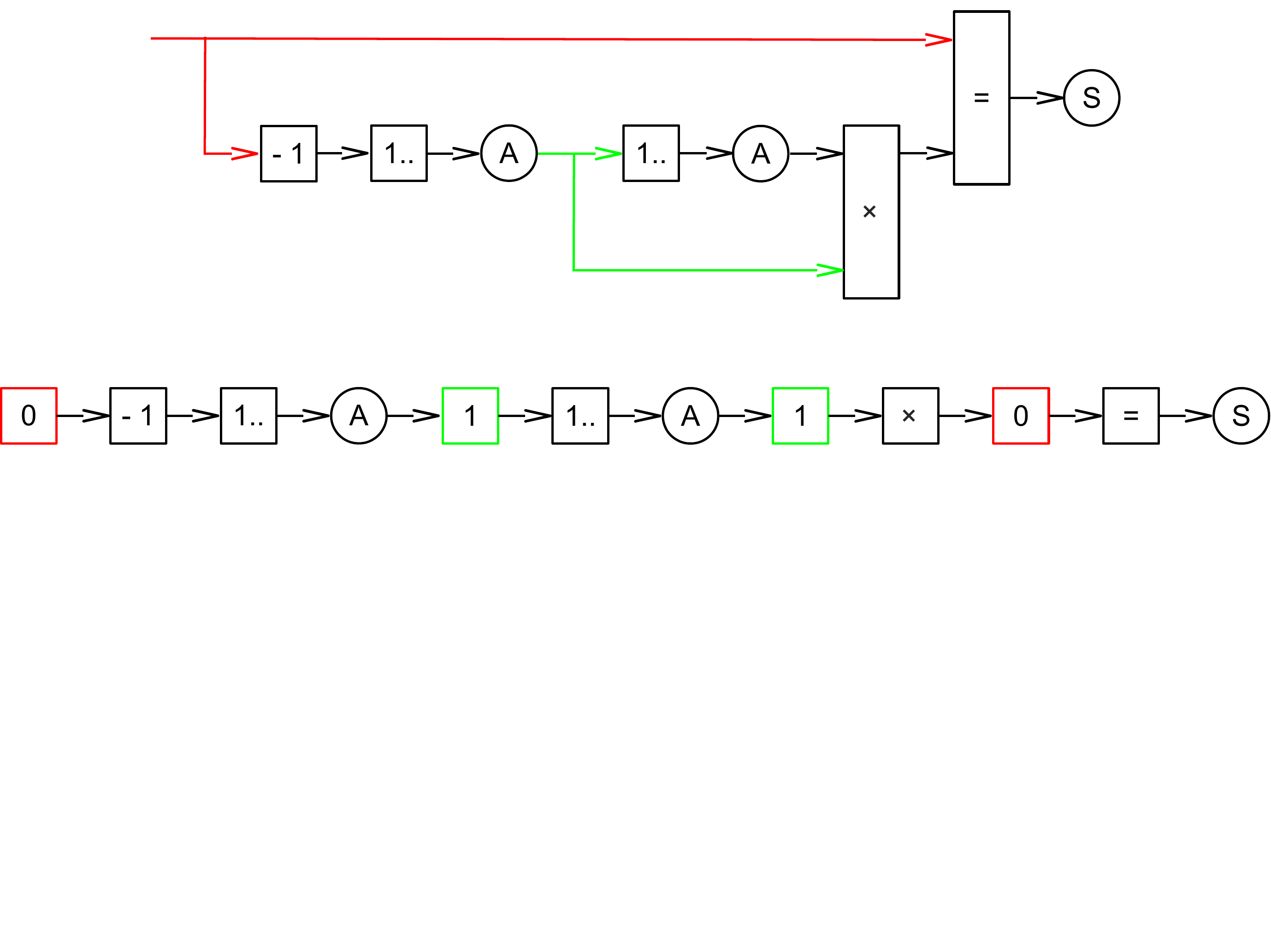}
\caption{{\it Non-deterministic dataflow graph} for deciding primality (top).
Dataflow graph compiled into a sequence of {\it non-deterministic combinators} (bottom).}
\label{skynet}
\end{center}
\end{figure}

Finally, the desugared comprehension contains functions like $(- \: 1)$,  $(\times)$ and $(=)$
that map scalars to scalars, yet we need functions that map sets to superpositions of sets.
Fortunately, sensible lifted forms for these functions are easily defined.
For example
\begin{eqnarray*}
{\rm pred}\:' & = &  {\rm unit}_A \cdot ({\rm map}_S \: (- \: 1))\\
{\rm times}\:' \; x\:' \; y\:' & = &  {\rm unit}_A \; \{ \; x \times y \; | \; x \in x\:', \; y \in y\:' \; \}\\
 {\rm equals}\:' \; x\:' \; y\:' & = & {\rm unit}_A \; \{ \; x \; | \; x \in x\:', \; y \in y\:', \; x = y \; \}
\end{eqnarray*}
\noindent where $x\:'$ and $y\:'$ are of type $\{a\}$.
Using these lifted functions and those defined previously,  the non-deterministic comprehension for
deciding primality can be translated as follows:
\[
\lambda n\:' \shortrightarrow ({\rm unit}_A \: n\:' \bindA {\rm pred}\:' \bindA \iota' \bindA {\rm amb}' \bindA 
\]
\[
\lambda x\:' \shortrightarrow ({\rm unit}_A \: x\:' \bindA \iota'  \bindA {\rm amb}' \bindA 
\]
\[
({\rm times}\:' \; x\:') \bindA ({\rm equals}\:' \; n\:') \bindA {\rm some}\:'))
\]
\noindent where $n\:'$ is of type $\{ a \}$. 
This was a lot of work, but we have reaped a tangible benefit,
namely, non-deterministic comprehensions can now be rendered as {\it dataflow graphs}. 
In Figure \ref{skynet} (top) boxes with one input have type signatures matching $f\:'$ and boxes with two
inputs have type signatures matching $g\:'$.
Arrows connecting pairs of boxes are instances of $(\bind_{\!\!\!\!A})$.
Junctions correspond to values of common subexpressions bound to variable names introduced by $\lambda$--expressions.
Lastly,  $\mycirc{\rm A}$ is $amb\:'$ and $\mycirc{\rm S}$ is $some\:'$.
This result is important because, without the amenity (provided by all general purpose programming languages) 
of being able to define and name functions, 
comprehension syntax quickly becomes unwieldy.
For this reason, we make extensive use of 
dataflow graphs as a visual programming language in the remainder of this paper.

\section {From Dataflow Graphs to Combinators}

One might assume that evaluation of dataflow graphs containing junctions would require an interpreter
with the ability to create and apply anonymous functions or {\it closures}.
These would contain the environments needed to lookup the values bound to variable names introduced by $\lambda$--expressions.
Happily, this turns out to be unnecessary.
In this section we show how dataflow graphs can be evaluated by a stack machine and
define a set of combinators that can be used to construct stack machine programs.

In general, combinators apply functions to one (or two) values of type $\{a\}$ popped from the front of the stack
and then push a result of type $\{a\}$ back onto the stack.
Since dataflow graphs are non-deterministic, the stack machine is also. 
This means that each combinator $f\:''$ transforms a stack of sets into a superposition of stacks of sets
\[
\shortrightarrow \fbox{\phantomas{$g\:'$}{$f\:''$}} \shortrightarrow \;\;\; :: \;\;\; \stackl \setl a \setr \stackr \shortrightarrow \supl \stackl \setl a \setr \stackr \supr.
\]
Unary operators $f\:'$ can be converted to combinators of type $f\:''$ as follows:
\[
f\:'' \; \; (x\:':s \:'') = {\rm map}_A \; (: \: s\:'') \; (f\:' \; x\:')
\]
\noindent where stack $s\:''$ is of type $[ \{ a \} ]$, ${\it map}_A$ maps functions over superpositions
and $(: \: s\:'')$ is the function that pushes sets onto the front of $s\:''$.
Note that $f\:''$ does not change the length of the stack;
it consumes one value and leaves one value behind.
Binary operators $g\:'$  can also be converted to combinators of type $f\:''$ as follows:
\[
g\:'' \; \; (x\:' : y\:' : s\:'') = {\rm map}_A \; (: \: s\:'') \; (g\:' \; x\:' \; y \:').
\]
\noindent Note that $g\:''$ decreases the length of the stack by one;
it consumes two values and leaves one value behind.
The combinator forms of $some\:'$ and $none\:'$ are slightly different;
they do not push a result onto the stack.
Instead, they pop the stack when a non-deterministic computation has yielded a satisfactory intermediate result
(whether that is something or nothing) and fail otherwise:
\[
{\rm some}\:'' \; (x\:':s\:'') = 
\left\{\begin{array}{ll}
\fail & {\rm if}\;\;x\:' = \nil\\
{\rm unit}_A \; s\:'' & {\rm otherwise}
\end{array}
\right.\]
\[
{\rm none}\:'' \; (x\:':s\:'') = 
\left\{\begin{array}{ll}
{\rm unit}_A \; s\:'' & {\rm if}\;\;x\:' = \nil\\
\fail & {\rm otherwise.}
\end{array}
\right.\]



\noindent Multiple functions can be applied to a single value by pushing copies of the value
onto the top of the stack and then applying the functions to the copies.
This preserves the value for future use and eliminates the need for closures.
Accordingly, we define a set of combinators that copy and push values located
at different positions within the stack
\[
{\rm x}\:''_k \;  (s\:'' ) \; = {\rm unit}_A \; ((s\:'' \; {\rm !!}  \;(n- k)) : s\:'')
\]
\noindent where $k \in \{0..9\}$, $(!!)$ returns the element of a list with a given index, and $n$ is the length of $s\:''$.
With this last puzzle piece in place, we can finally do what we set out to do, namely,
compile the comprehension for deciding primality into a sequence of combinators

{\small
\[
{\rm x}\:''_0 \kleisliA {\rm pred}\:'' \kleisliA  \iota\:'' \kleisliA {\rm amb}\:'' \kleisliA  {\rm x}\:''_1 \kleisliA  \iota\:'' \kleisliA  {\rm amb}\:''  
\]
\vspace{-0.125in}
\[
\kleisliA  {\rm x}\:''_1 \kleisliA {\rm times}\:'' \kleisliA {\rm x}\:''_0 \kleisliA {\rm equals}\:'' \kleisliA {\rm some}\:''
\]
}
\noindent where $(\kleisli)$ is {\it Kleisli composition}
\[
f \kleisli g = (\bind g) \cdot f
\]
In Figure \ref{skynet} (bottom) boxes are functions with type signatures matching $f\:''$.
Arrows connecting pairs of boxes are instances of $(\kleisli_{\!A})$.
Lastly,  $\mycirc{\rm A}$ is $amb\:''$ and $\mycirc{\rm S}$ is $some\:''$.

\newcommand*{\mybox}[1]{%
  \funbox{\raisebox{0pt}[0.4\baselineskip][0.05\baselineskip]{%
    #1}}}

\section{Reified Actor Comprehensions}


The last two sections of the paper demonstrated that:
1) Non-deterministic comprehensions can be represented as dataflow graphs; and
2) Dataflow graphs can be compiled into sequences of combinators that evaluate comprehensions
by transforming the state of an abstract machine.
In this section we describe a visual programming language for specifying behaviors manifested by reified actors
in a virtual world.
All results from prior sections apply.
However, non-determinism must be combined with other effects to construct a monad more general than $A$ which we call $R$
 (for {\it reified actor}).
In addition to representing superpositions, monad $R$ provides mutation of a threaded global state and
data logging so that behaviors composed of combinators can report the time they consume.
The boxes of dataflow graphs with one and two inputs now have types
$\setl Actor \setr \shortrightarrow \supl \setl Actor \setr \supr\:'$ and
$\setl Actor \setr \shortrightarrow \setl Actor \setr \shortrightarrow \supl \setl Actor \setr \supr\:'$
where $\supl \: \supr\:'$ is the type constructor of monad $R$.
Arrows connecting boxes are instances of $(\bind_{\!R})$.
Combinators now have type $[\setl Actor \setr] \shortrightarrow \supl [\setl Actor \setr] \supr\:'$
and are composed with $(\kleisli_{\!R})$.

Combinators can be divided into the categories: {\it generators}, {\it guards}, {\it relations}, and {\it actions}.
Generators are unary operators that characterize sets of actors using
the devices of groups, containment, bonds, and neighborhood (Table \ref{table:generators}).
They can be composed to address different sets.
For example, an actor's siblings can all be
addressed using the subgraph  \mybox{$\phantomas{A}{\hat{}}$} $\rightarrow$ \mybox{$\phantomas{A}{@}$} .
\noindent Generators can also be composed with guards (Table \ref{table:guards}).
This can be used either to address single actors or to specify preconditions for actions.
For example, the subgraph
\mybox{$\phantomas{A}{\hat{}}$} $\rightarrow$ \mybox{$\phantomas{A}{@}$} $\rightarrow \mycirc{\rm A}$
addresses a single sibling while the subgraph 
\mybox{$\phantomas{A}{\#}$} $\rightarrow \mycirc{\rm N}$
fails if the actor has a neighbor.

{\renewcommand{\arraystretch}{1.25} 
\begin{small}
\begin{table}[ht]
\caption{Unary generators.}
\centering
\begin{tabular}{| c | c | l |}
\hline
Name & Abbrev. & Definition\\
\hline
$\chands$ & $|$ & actor sharing hand with $x$\\
\hline
$\cnexts$ & $>$ & actor with directed bond from $x$\\
\hline
$\productrevs$ & $<$ & actor with directed bond to $x$\\
\hline
$\cbonds$ & $:$ & union of hands, nexts and prevs\\
\hline
$\cneighbors$ & \# & actors in neighborhood of $x$\\
\hline
$\ccontents$ & @ & actors that are contained in $x$\\
\hline
$\productarents$ & ${\large \hat{}}$ & actor that contains $x$\\
\hline
$\cmembers$ & * & members of group of $x$\\
\hline
$\cothers$ & + & members of group of $x$ but not $x$\\
\hline
\end{tabular}
\label{table:generators}
\end{table}
\end{small}
}

{\renewcommand{\arraystretch}{1.25} 
\begin{small}
\begin{table}[ht]
\caption{Unary guards.}
\centering
\begin{tabular}{| c | c | l |}
\hline
Name & Abbrev. & Definition\\
\hline
${\rm amb}$ & A & non-deterministic choice\\
\hline
${\rm some}$ & S & Fail if empty.\\
\hline
${\rm none}$ & N & Fail if non-empty.\\
\hline
\end{tabular}
\label{table:guards}
\end{table}
\end{small}
}

Relations exist for testing equality and type equivalence (Table \ref{table:relations}).
They are binary operators and are generally applied to singleton sets in combination with guards to specify preconditions for actions.
When applied to non-singleton sets, the equality operator and its negation
compute set intersection and difference.

{\renewcommand{\arraystretch}{1.25} 
\begin{small}
\begin{table}[ht]
\caption{Binary relations.}
\centering
\begin{tabular}{| c | c | l |}
\hline
Name & Abbrev. & Definition\\
\hline
$\csame$ & $ = $ & set intersection\\
\hline
$\cdifferent$ & $! =$ & set difference\\
\hline
$\csimilar$ & $\sim$ & all $x$ type equivalent to some $y$\\
\hline
$\cdissimilar$ & $!\sim$ & all $x$ type equivalent to no $y$\\
\hline
\end{tabular}
\label{table:relations}
\end{table}
\end{small}
}


Actions for modifying actors' persistent states are the final category of boxes in dataflow graphs.
Actions are rendered as grey boxes and are executed only after all non-actions
have been evaluated and only if no guard has failed.
All actions are reversible but the masses and types of primitive combinators and empty objects are immutable.
The full set of unary and binary actions is shown in Tables \ref{table:unary} and \ref{table:binary}.

{\renewcommand{\arraystretch}{1.25} 
\begin{small}
\begin{table}[ht]
\caption{Unary actions.}
\centering
\begin{tabular}{| c | c | l |}
\hline
Name & Abbrev. & Definition\\
\hline
$\cdrop$ & $! \; |$ & Delete hand of $x$.\\
\hline
$\ckillnext$ & $! >$ & Delete directed bond from $x$.\\
\hline
$\:\;\ckillprev$ & $! <$ & Delete directed bond to $x$.\\
\hline
$\cquit$ &  $\quit$ & Remove $x$ from its group.\\
\hline
$\cexpel$ & $\expel$ & Place $x$ inside its parent's parent.\\
\hline
$\cdigest$ & $\digest$ & Reduce $x$ to primitive combinators.\\
\hline
$\con$ & $/$ & Replace combinator with behavior.\\
\hline
$\coff$ & \textbackslash & Replace behavior with combinator.\\
\hline
\end{tabular}
\label{table:unary}
\end{table}
\end{small}
}

{\renewcommand{\arraystretch}{1.25} 
\begin{small}
\begin{table}[ht]
\caption{Binary actions.}
\centering
\begin{tabular}{| c | c | l |}
\hline
Name & Abbrev. & Definition\\
\hline
$\cgrab$ & $|$ & Create hand between $x$ and $y$.\\
\hline
$\cmakenext$ & $>$ & Create directed bond from $x$ to $y$.\\
\hline
$\:\;\cmakeprev$ & $<$ & Create directed bond from $y$ to $x$.\\
\hline
$\cjoin$ & $\join$ & $x$ joins group of $y$.\\
\hline
$\ceat$ &  $\eat$ & Place $x$ inside $y$.\\
\hline
$\ccompose$ & $\kleisli$ & Replace $x$ with $x \kleisli_R y$.\\
\hline
$\cswap$ & $\%$ & $x$ and $y$ swap positions and bonds.\\
\hline
\end{tabular}
\label{table:binary}
\end{table}
\end{small}
}

Where data dependencies determine order of execution, this order is followed.
Where it would otherwise be underdetermined, two devices are introduced to specify execution order.
First, all actions return their first (or only) argument if they succeed.
This allows one action to provide the input to a second and (when employed) 
introduces a data dependency that determines execution order.
Second, execution order can be explicitly specified using dotted {\it control lines}.

In addition to non-determinism and mutable threaded state, instances of monad $R$ also possess a data logging
ability that is used to instrument combinators so that behaviors comprised of them can report the time they consume.
Because the unit of time is one primitive operation of the abstract machine, 
most primitive combinators increase logged time by one when they are run.
Significantly, this occurs on all branches of the non-deterministic computation until a branch succeeds
so that the full cost of simulating non-determinism on a (presumed) deterministic substrate
by means of backtracking is accounted for.
Two kinds of combinators increase logged time by amounts other than one.
Since the time required to compute set intersections and differences is the product of the sets' lengths,
for binary relations, the logged time is increased by this value instead
(which equals one in the most common case of singleton sets).
Finally, actions that change the position of an actor, {\it e.g.,} {\it join}, pay an additional
time penalty proportional to the product of the actor's mass and the $L_1$ distance moved.

Ideally, the actor model described in this paper would be reified as an ACA so that self-replicating
programs consume real physical resources.
Actors in an embedded group might share a single processor or might jointly occupy a 2D area of fixed
size that collects a fixed amount of light energy per unit time.
The effect would be the same; the number of primitive abstract machine operations executed
per unit time by the processor (or in the area) would be fixed.

For the time being, we implement the reified actor model as an event driven simulation using a priority queue \citep{gillespie}.
Event times are modeled as Poisson processes associated with embedded groups and event rates
are consistent with the joint consumption by actors in groups of  finite time resources.
Events are of two types.
When a {\it diffusion event} is at the front of the queue, the position of the group in its neighborhood is randomly changed
(as previously described).
Afterwards, a new diffusion event associated with the same group is enqueued.
The time of the new event is a sample from a distribution with density
$f_D(t) = D \: e^{- D \: t / (m \: s)}  / (m \: s)$ where $m$ is mass, $s$ is distance, and $D$
is the ratio of the time needed to execute one primitive operation and 
the time needed to transport a unit mass a unit distance.
As such, it defines the relative cost of computation and data transport in the ACA substrate.\footnote{In all of our experiments $D$ equals 10.}

When an {\it action event} is at the front of the queue, a behavior is chosen at random
from among all actors of type behavior in the group.
After the behavior is executed, the time assigned to the new action event is a sample from a distribution
with density $f_A(t) = e^{-t / c}  / c$ where $c$ is the time consumed by the behavior.

\begin{figure}[t]
\begin{center}
\includegraphics[scale = 0.32, trim = 20 20 0 0]{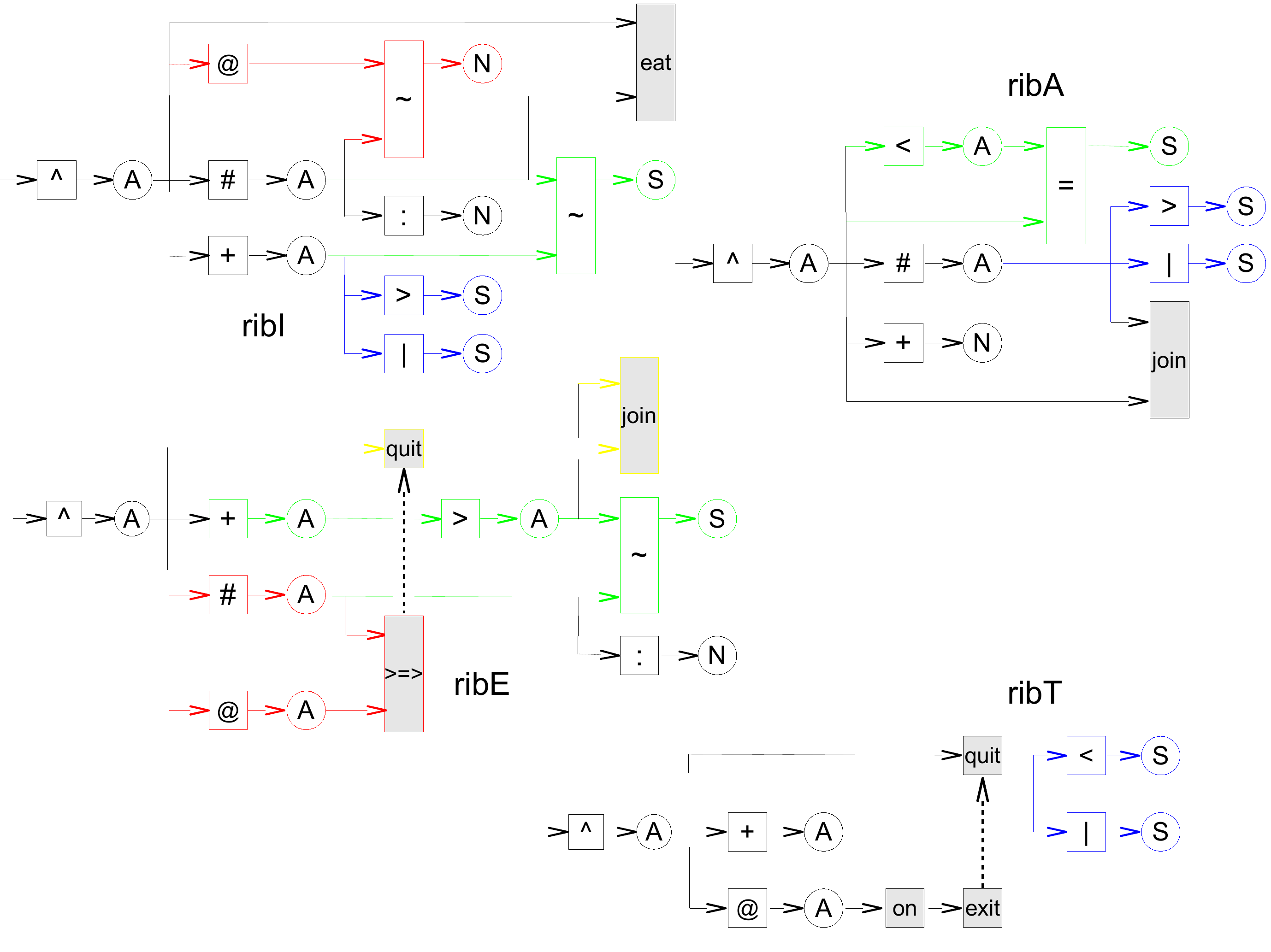}
\caption{Behaviors defining a ribosome.}
\label{ribAIED}
\end{center}
\end{figure}

\section{Computational Ribosomes}

Biological enzymes can be reified as chains of nucleotides or amino acids.
The first can be read and copied but are spatially distributed and purely representational; 
the second are representationally opaque but compact and metabolically active.
Comprehensions can be compiled into sequences of primitive combinators and reified in analogous ways.
A {\it plasmid} is a compiled comprehension reified as a chain of actors of type combinator linked with directed bonds:
\noindent 
\begin{eqnarray*}
P = \lreifiedobj c_0 \rreifiedobj^- \nextbond  \lreifiedobj c_1 \rreifiedobj^- \nextbond \cdots \nextbond \lreifiedobj c_{N-1} \rreifiedobj^-
\end{eqnarray*}
\noindent where $(>)$ is a directed bond and $\lreifiedobj \; \rreifiedobj$ denotes an actor that is reified at the root level.
A single undirected bond (not shown) closes the chain and marks the plasmid's {\it origin}.
While plasmids are spatially distributed chains of many actors, {\it enzymes} are single actors of type behavior:
\begin{eqnarray*}
E = \lreifiedobj c_0 \kleisli_R \;  c_1 \kleisli_R \cdots \kleisli_R c_{N-1} \: \rreifiedobj^+.
\end{eqnarray*}

Biological ribosomes are arguably the most important component of the fundamental dogma \citep{watson}.
They translate messenger RNA into polypeptides using a four stage process of
{\it association}, {\it initiation}, {\it elongation} and {\it termination}.
We can construct a {\it computational ribosome} that will translate plasmids into enzymes
by defining four behaviors with analogous functions  (Figure \ref{ribAIED}),
reifying the behaviors as enzymes, and placing them inside an actor of type object
\[
R = \lreifiedobj  E_{\rm ribA}, \: E_{\rm ribI}, \: E_ {\rm ribE}, \: E_ {\rm ribT} \: \rreifiedobj_0.
\]

Behavior {\it ribA} first checks to see if $R$ possesses a self-directed bond.\footnote{Ribosomes without this bond are disabled
and serve solely as {\it models} for {\it factories}, {\it i.e.,} as compositional information.}
If so, {\it ribA} attaches $R$ to the plasmid by adding it to  the group of the initial combinator, $\lreifiedobj c_0 \rreifiedobj^-.$
Next, {\it ribI} finds an actor in the neighborhood with type matching
$\lreifiedobj c_0 \rreifiedobj^-$ and places it inside $R$.
When $R$ is at position $n$ on the plasmid, {\it ribE} finds a neighbor with type matching $\lreifiedobj c_{n+1} \rreifiedobj^-$
and composes it with the combinator inside $R$, {\it i.e.,} with $[ c_0 \kleisli_R \cdots \kleisli_R c_n ]^-$.
It then advances the position of $R$ to $n+1$.
This process continues until $R$ reaches $N-1$,  
at which point {\it ribT} promotes the combinator to a behavior, 
expels it, and detaches $R$ from the plasmid.

If a ribosome and a plasmid are placed in the world with a supply of primitive combinators,
the ribosome manufactures the enzyme described by the plasmid
\[{\textstyle
R + P_{\:b}+ \sum_{\: C} m_b( c ) \: \lreifiedobj c \rreifiedobj^- \rightarrow R + P_{\:b} + E_{b}
}\]
\noindent where $C$ is the set of 42 primitive combinators and $m_b( c )$ is the number of combinators
of type $c$ in $P_{\:b}$ and $E_{b}$, {\it i.e.,} the plasmid and enzyme reifications of behavior $b$.

\begin{figure}[t]
\begin{center}
\includegraphics[scale = 0.32, trim = 20 40 0 0]{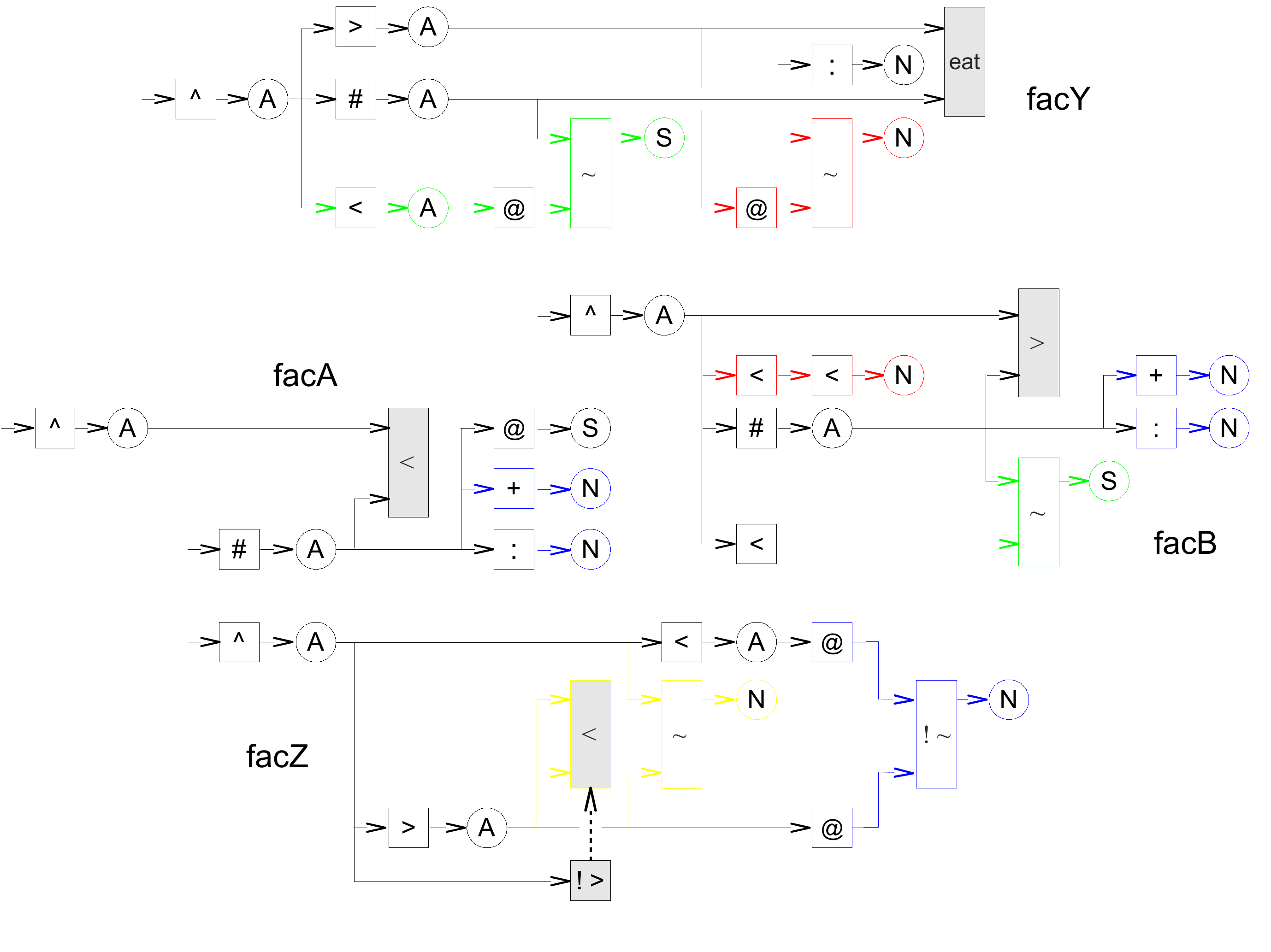}
\caption{Behaviors defining a factory.}
\label{facABYZ}
\end{center}
\end{figure}

Now that we have a ribosome, we need something to do with it.
We could (of course) use ribosomes to synthesize the enzymes of which they themselves are comprised.
However, it would be more interesting if these enzymes were then used to construct additional ribosomes.
To accomplish this, we need a `machine' that will collect the finished enzymes and place them inside an object of the correct type.
We call this machine a {\it factory}.
Factories are copiers of {\it compositional information}, which is heritable information distinct from the {\it genetic information}
that ribosomes translate into enzymes.
A factory can be constructed by reifying the behaviors defined in Figure \ref{facABYZ} as enzymes
and placing them inside an object with a type distinct from that of ribosomes: 
\[
F = \lreifiedobj  E_{\rm facA}, \: E_{\rm facB}, \: E_{\rm facY}, \: E_{\rm facZ}, \: E_{\rm facZ\:'} \rreifiedobj_1.
\]
\noindent Behavior {\it facA} creates a directed bond with any unbonded non-empty object it finds in the factory's neighborhood.
This object and its contents serve as the {\it model}.
Behavior {\it facB} creates a second directed bond from the factory to an empty object with type matching the model.
This object serves as the container for the {\it product.}
Behavior {\it facY} moves behaviors from the neighborhood similar to those in the model into the product.
Behavior {\it facZ} recognizes when the product contains the full set of behaviors and deletes the bond connecting it to the factory.
Behavior ${\it facZ}\:'$ does the same but also installs a self-directed bond on ribosomes 
that enables their association behaviors
(elements unique to ${\it facZ}\:'$ are yellow in Figure \ref{facABYZ}).
\newcommand{\obj} {\lreifiedobj \: \rreifiedobj}

\begin{figure}[t]
\begin{center}
\includegraphics[scale = 0.3, trim = 30 0 0 0]{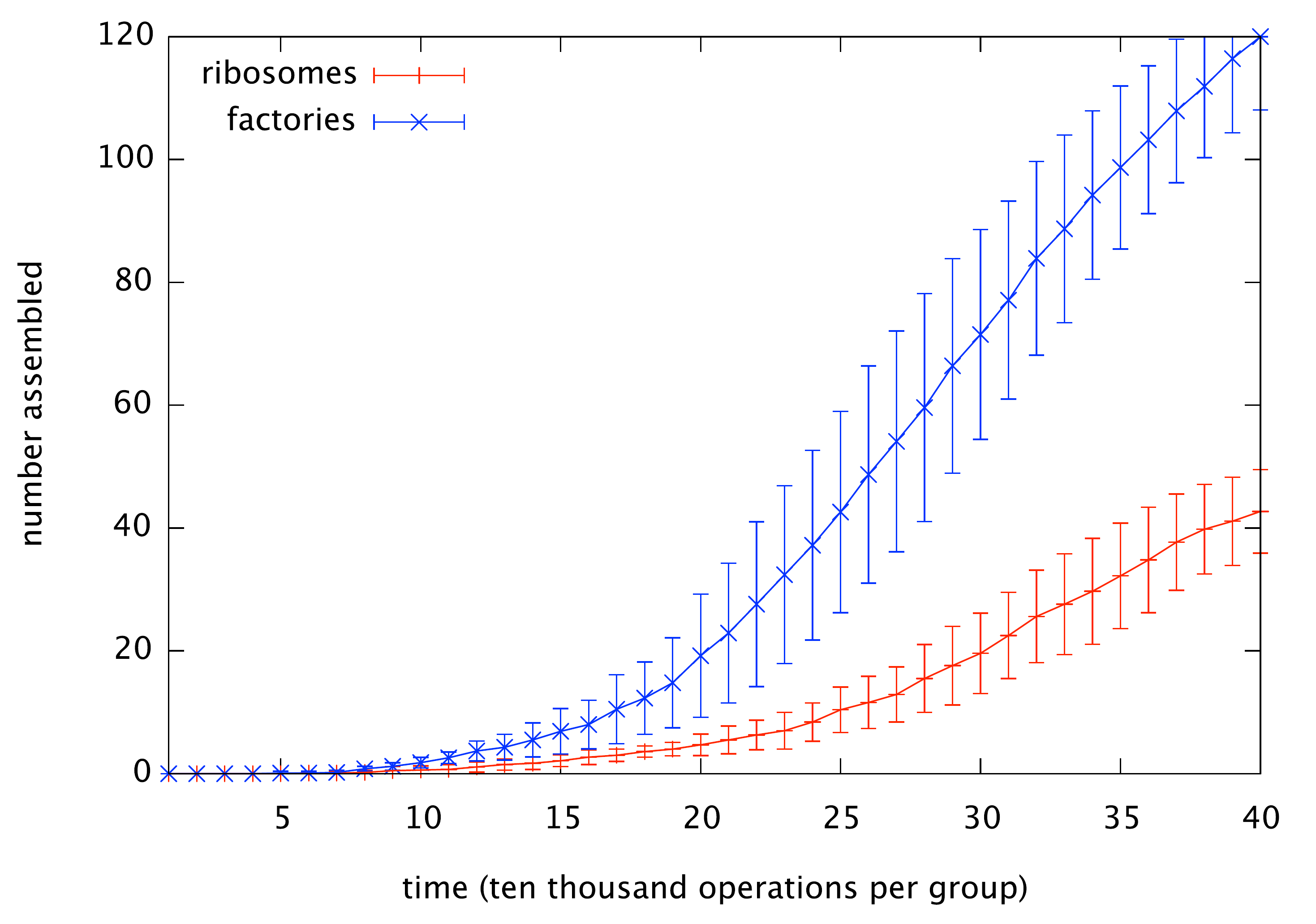}
\caption{Average increase in numbers of ribosomes and factories (ten runs).  Error bars show $\pm$ one standard deviation.}
\label{symbiosis}
\end{center}
\end{figure}

As an initial experiment, we demonstrate mutual replication of a mixed population of ribosomes and factories.
Plasmids $P_{\:b}$ encoding enzymes $E_b$ comprising ribosomes and factories
are placed in a 2D virtual world consisting of $64 \times 64$ sites together with a large surplus of
ribosomes ($r = 64$) and single instances of factories with ribosome and factory models, 
$F_R$ and $F_F$.
The supply of primitive combinators and empty objects is replenished as instances are incorporated
into enzymes and products.
Consequently, the concentration of consumables is held constant.
Plasmids and consumables required for synthesis of factory enzymes
are overrepresented relative to those for ribosomal enzymes:
\[{\textstyle
r \: R + F_R + F_F + \sum_{\:B} P_{\:b} + 2 \obj_0 + 3 \obj_1 + \sum_{\:B}  \sum_{\:C} m_b ( c ) \lreifiedobj c \rreifiedobj^-
}\]
\[{\textstyle
\rightarrow (r + 1) \: R +  2 \: F_R + 2 \: F_F + \sum_{\:B} P_{\:b}
}\]
\noindent where the multiset $B = \{ \; b \;\; | \;\; E_b \; \in \; 2 \: R \; \cup \; 3 \: F \:\}$.
We observe that the ribosomes synthesize the enzymes encoded by the plasmids
and these are then used by the factories to construct additional ribosomes and factories.
See Figure \ref{symbiosis}.


\section{Conclusion}


Fifty years after von Neumann described his automaton, it remains a paragon of non-biological life.
The rules governing CAs are simple and physical, and partly for this reason, the automaton von Neumann constructed using them
is uniquely impressive in its semantic closure.
Yet perhaps because RASPs are (in comparison with CAs) relatively well-appointed hosts,
self-replicating programs in conventional programming languages seem somehow less convincing.
All self-replicating programs must lift themselves up by their own bootstraps, yet not all programs lift themselves the same distance.
The field of programming languages has made remarkable advances in the years since von Neumann conceived his automaton.
Modern functional programming languages like Haskell bear little resemblance to the
machine languages that are native to RASPs.
In this paper, we have attempted to show that programs defined using seemingly exotic constructs like non-deterministic
comprehensions can in fact be compiled into sequences of combinators with simple, 
well-defined semantics.
Moreover, because they do not have address operands, these combinators can be reified in a virtual world
as actors of only a few fixed types.
This makes it possible to build programs that build programs from components delivered by diffusion using processes
that resemble chemistry as much as computation.

\section{Acknowledgements}
Special thanks to Joe Collard.  Thanks also to Dave Ackley, Stephen Harding, Barry McMullin and Darko Stefanovic.
\small
%
\bibliographystyle{apalike}
\bibliography{ecal15}  

\begin{thebibliography}{}

\bibitem[Ackley, 2013]{bespoke}
Ackley, D. (2013).
\newblock Bespoke physics for living technology.
\newblock {\em Artificial Life}, 34:381--392.

\bibitem[Adami et~al., 1994]{adami1}
Adami, C., Brown, C.~T., and Kellogg, W. (1994).
\newblock Evolutionary learning in the 2{D} artificial life system ``{A}vida".
\newblock In {\em Artificial Life IV}, pages 377--381. MIT Press.

\bibitem[Berman and Simon, 1988]{berman}
Berman, P. and Simon, J. (1988).
\newblock Investigations of fault-tolerant networks of computers.
\newblock In {\em STOC}, pages 66--77.

\bibitem[Berry and Boudol, 1990]{cham}
Berry, G. and Boudol, G. (1990).
\newblock The chemical abstract machine.
\newblock In {\em Proceedings of the 17th ACM SIGPLAN-SIGACT Symposium on
  Principles of Programming Languages}, POPL '90, pages 81--94, New York, NY,
  USA. ACM.

\bibitem[Felleisen, 1990]{felleisen}
Felleisen, M. (1990).
\newblock On the expressive power of programming languages.
\newblock In {\em ESOP'90}, pages 134--151. Springer.

\bibitem[Fontana and Buss, 1999]{fontana}
Fontana, W. and Buss, L.~W. (1999).
\newblock What would be conserved if the tape were played twice?
\newblock In Cowan, G.~A., Pines, D., and Meltzer, D., editors, {\em
  Complexity}, pages 223--244. Perseus Books, Cambridge, MA, USA.

\bibitem[Gillespie, 1977]{gillespie}
Gillespie, D.~T. (1977).
\newblock Exact stochastic simulation of coupled chemical reactions.
\newblock {\em The Journal of Physical Chemistry}, 81(25):2340--2361.

\bibitem[Hutton, 2004]{hutton}
Hutton, T.~J. (2004).
\newblock A functional self-reproducing cell in a two-dimensional artificial
  chemistry.
\newblock In {\em Proc. of the 9th Intl. Conf. on the Simulation and Synthesis
  of Living Systems (ALIFE9)}, pages 444--449.

\bibitem[Laing, 1977]{laing}
Laing, R.~A. (1977).
\newblock Automaton models of reproduction by self-inspection.
\newblock {\em Journal of Theoretical Biology}, 66(1):437--456.

\bibitem[Mc{C}arthy, 1963]{mccarthy}
Mc{C}arthy, J. (1963).
\newblock A basis for a mathematical theory of computation.
\newblock In {\em Computer Programming and Formal Systems}, pages 33--70.
  North-Holland.

\bibitem[Nakamura, 1974]{nakamura}
Nakamura, K. (1974).
\newblock Asynchronous cellular automata and their computational ability.
\newblock {\em System Comput. Controls}, 15(5):56--66.

\bibitem[Nehaniv, 2004]{nehaniv}
Nehaniv, C.~L. (2004).
\newblock Asynchronous automata networks can emulate any synchronous automata
  network.
\newblock {\em IJAC}, 14(5-6):719--739.

\bibitem[Pattee, 1995]{pattee}
Pattee, H. (1995).
\newblock Evolving self-reference: Matter, symbols, and semantic closure.
\newblock {\em Communication and Cognition - Artificial Intelligence},
  12:9--27.

\bibitem[Paun, 1998]{paun}
Paun, G. (1998).
\newblock Computing with membranes.
\newblock {\em Journal of Computer and System Sciences}, 61:108--143.

\bibitem[Ray, 1994]{ray}
Ray, T.~S. (1994).
\newblock An evolutionary approach to synthetic biology, {Z}en and the art of
  creating life.
\newblock {\em Artificial Life}, 1:179--209.

\bibitem[Smith et~al., 2003]{turney}
Smith, A., Turney, P.~D., and Ewaschuk, R. (2003).
\newblock Self-replicating machines in continuous space with virtual physics.
\newblock {\em Artificial Life}, 9(1):21--40.

\bibitem[Wadler, 1990]{wadler92}
Wadler, P. (1990).
\newblock Comprehending monads.
\newblock In {\em Proceedings of the 1990 ACM Conference on LISP and Functional
  Programming}, LFP '90, pages 61--78, New York, NY, USA. ACM.

\bibitem[Watson and Crick, 1953]{watson}
Watson, J.~D. and Crick, F.~H. (1953).
\newblock Molecular structure of nucleic acids.
\newblock {\em Nature}, 171(4356):737--738.

\bibitem[Williams, 2014]{williams14}
Williams, L. (2014).
\newblock Self-replicating distributed virtual machines.
\newblock In {\em Proc. of the 14th Intl. Conf. on the Simulation and Synthesis
  of Living Systems (ALIFE14)}, pages 711--718.

\end{thebibliography}

\end{document}